\documentclass[runningheads]{llncs}
\usepackage{graphicx}
\usepackage{caption}
\usepackage[pagebackref,breaklinks,colorlinks]{hyperref}
\usepackage{tikz}
\usepackage{comment}
\usepackage{amsmath,amssymb} 
\usepackage{color}
\usepackage{bbding}
\usepackage{pifont}
\usepackage[capitalize]{cleveref}
\usepackage{adjustbox}
\usepackage{color}
\usepackage{subfigure}
\usepackage{booktabs}

\usepackage[accsupp]{axessibility}  
\usepackage{lipsum}
\usepackage{appendix}
\usepackage[misc]{ifsym}

\makeatletter
\newcommand{\printfnsymbol}[1]{%
        \textsuperscript{\@fnsymbol{#1}}%
}
\makeatother

\crefname{section}{Sec.}{Secs.}
\Crefname{section}{Section}{Sections}
\Crefname{table}{Table}{Tables}
\crefname{table}{Tab.}{Tabs.}

\begin{document}
\pagestyle{headings}
\mainmatter

\title{KD-MVS: Knowledge Distillation Based Self-supervised Learning for Multi-view Stereo} 

\titlerunning{KD-MVS: Knowledge Distillation Based Self-supervised Learning for MVS}

\renewcommand{\thefootnote}{\fnsymbol{footnote}}
\footnotetext{This work is done by the first four authors as interns at Megvii Research.}

\author{Yikang Ding\textsuperscript{\rm 1,}\textsuperscript{\rm 2}
\,\,
Qingtian Zhu\textsuperscript{\rm 1}
\,\,
Xiangyue Liu\textsuperscript{\rm 1}
\,\,
Wentao Yuan\textsuperscript{\rm 1}
\,\, \\
Haotian Zhang\textsuperscript{\rm 1}\thanks{Corresponding author (zhanghaotian@megvii.com).}
\,\,
Chi Zhang\textsuperscript{\rm 1}
}
\institute{$^1$Megvii Research \quad $^2$Tsinghua University}

\authorrunning{Y. Ding et al.}

\maketitle

\begin{abstract}

Supervised multi-view stereo (MVS) methods have achieved remarkable progress in terms of reconstruction quality, but suffer from the challenge of collecting large-scale ground-truth depth. In this paper, we propose a novel self-supervised training pipeline for MVS based on knowledge distillation, termed \textit{KD-MVS}, which mainly consists of self-supervised teacher training and distillation-based student training. Specifically, the teacher model is trained in a self-supervised fashion using both photometric and featuremetric consistency. Then we distill the knowledge of the teacher model to the student model through probabilistic knowledge transferring. With the supervision of validated knowledge, the student model is able to outperform its teacher by a large margin. Extensive experiments performed on multiple datasets show our method can even outperform supervised methods. Code is available at \href{https://github.com/megvii-research/KD-MVS}{https://github.com/megvii-research/KD-MVS}.

\end{abstract}


\section{Introduction}

\begin{figure}[t]
    \centering
    \includegraphics[width=\linewidth]{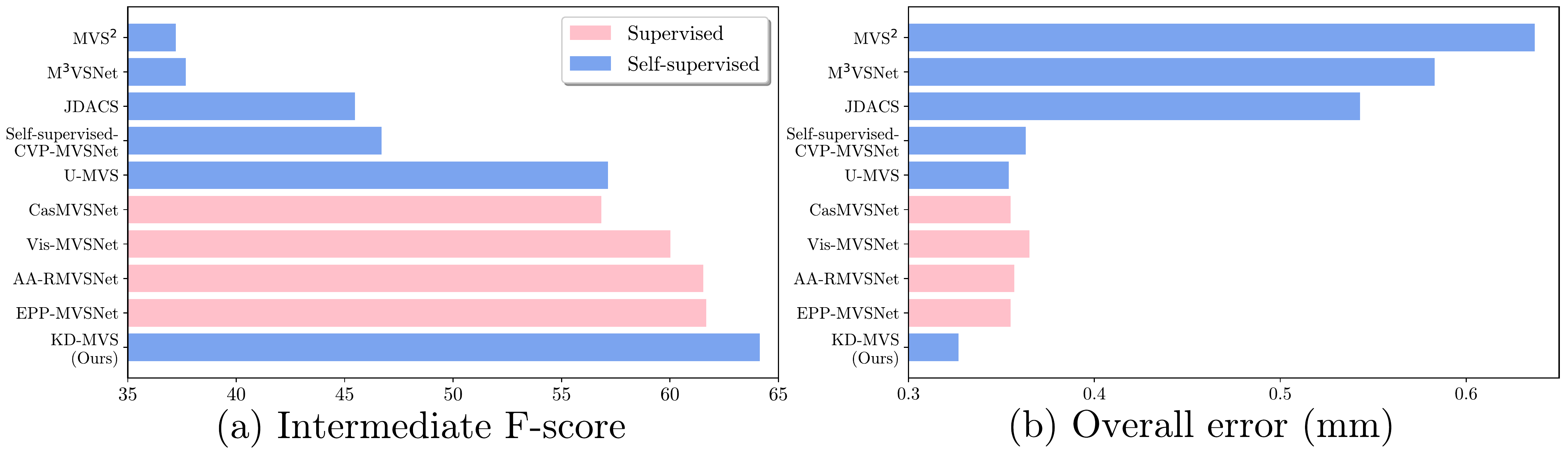}
    \caption{Visualized performance comparisons of state-of-the-art MVS methods on (a) Tanks and Temples benchmark~\cite{knapitsch2017tanks} and (b) DTU dataset~\cite{aanaes2016large}.}
    \label{fig:teaser}
\end{figure}

The task of multi-view stereo (MVS) is to reconstruct a dense 3D presentation of the observed scene using a series of calibrated images, which plays an important role in a variety of applications, e.g. augmented and virtual reality, robotics and computer graphics. Recently, learning-based MVS networks~\cite{yao2018mvsnet,yao2019recurrent,gu2020cascade,ding2021transmvsnet,ding2022enhancing,liao2022wt-mvsnet} have obtained impressive results.
However, supervised methods require dense depth annotations as explicit supervision, the acquisition of which is still an expensive challenge.
Subsequent attempts~\cite{khot2019Unsup-MVS,xu2021u-mvs,xu2021JDACS,yang2021selfsup-cvp,huang2020m3vsnet} have made efforts to train MVS networks in a self-supervised manner by using photometric consistency~\cite{khot2019Unsup-MVS,dai2019mvs2}, optical flow~\cite{xu2021u-mvs} or reconstructed 3D models~\cite{huang2020m3vsnet,yang2021selfsup-cvp}.

Though great improvement has been made, there is a significant gap in either reconstruction completeness or accuracy compared to supervised methods.

In this paper, we propose a novel self-supervised training pipeline for MVS based on knowledge distillation~\cite{hinton2015distilling}, named \textit{KD-MVS}. The pipeline of KD-MVS mainly consists of (a) self-supervised teacher training and (b) distillation-based student training.
In the self-supervised teacher training stage, the teacher model is trained by enforcing both the photometric consistency~\cite{khot2019Unsup-MVS} and featuremetric consistency between the reference view and the reconstructed views, which can be obtained via homography warping according to the estimated depth.
Unlike the existing self-supervised MVS methods~\cite{khot2019Unsup-MVS,yang2021selfsup-cvp,dai2019mvs2} that use only photometric consistency, we propose to use the internally extracted features to utilize the featuremetric consistency, which is different from the externally extracted features-based loss, e.g. perceptual loss~\cite{johnson2016perceptual-loss}. We analyze and show that the proposed internal featuremetric loss is more suitable for MVS and is able to help the self-supervised teacher model yield relatively complete and accurate depth maps.

The distillation-based student training stage consists of two main steps: the pseudo probabilistic knowledge generation and the student training. We first use the teacher model to infer raw depth maps on unlabeled training data and perform cross-view check to filter unreliable samples.
We then generate the pseudo probability distribution of the teacher model by probabilistic encoding.
The probabilistic knowledge can be transferred to the student model by forcing the predicted probability distribution of the student model to be similar to the pseudo probability distribution.
As a result, the student model can surpass its teacher and even outperform supervised methods.
Extensive experiments on DTU dataset~\cite{aanaes2016large}, Tanks and Temples benchmark~\cite{tnt} and BlendedMVS dataset~\cite{yao2020blendedmvs} show that KD-MVS brings significant improvement to off-the-shelf MVS networks, even outperforming supervised methods, as is shown in \cref{fig:teaser}.

Our main contributions are four-fold as follows:
\begin{enumerate}
\setlength{\itemsep}{0pt}
\setlength{\parsep}{0pt}
\setlength{\parskip}{0pt}
    \item[-] We propose a novel self-supervised training pipeline named KD-MVS based on knowledge distillation.
    \item[-] We design an internal featuremetric consistency loss to perform robust self-supervised training of the teacher model.
    \item[-] We propose to perform knowledge distillation to transfer validated knowledge from the self-supervised teacher to a student model for boosting performance.
    \item[-] Our method achieves state-of-the-art performance on Tanks and Temples benchmark~\cite{tnt}, DTU~\cite{aanaes2016large} dataset and BlendedMVS~\cite{yao2020blendedmvs} dataset.
\end{enumerate}


\section{Related Work}

\subsection{Learning-based MVS}
\subsubsection{Supervised MVS}
Learning-based methods for MVS have achieved impressive reconstruction quality. MVSNet~\cite{yao2018mvsnet} transforms the MVS task to a per-view depth estimation task and encodes camera parameters via differentiable homography to build 3D cost volumes, which will be regularized by a 3D CNN to obtain a probability volume for pixel-wise depth distribution. However, at cost volume regularization, 3D tensors occupy massive memory for processing.
To alleviate this problem, some attempts~\cite{yao2019recurrent,yan2020dense,wei2021aa} replace the 3D CNN by 2D CNNs and a RNN and some other methods~\cite{gu2020cascade,zhang2020visibility,cheng2020deep,yang2020cost} use a multi-stage approach and predict depth in a coarse-to-fine manner.

\subsubsection{Self-supervised MVS}
The key of self-supervised MVS methods is how to make use of prior multi-view information and transform the problem of depth prediction into other forms of problems. Unsup-MVS~\cite{khot2019Unsup-MVS} firstly handles MVS as an image reconstruction problem by warping pixels to neighboring views with estimated depth values.
Given multiple images, MVS$^2$~\cite{dai2019mvs2} predicts each view's depth simultaneously and trains the model using cross-view consistency. 
M$^3$VSNet~\cite{huang2020m3vsnet} makes use of the consistency between the surface normal and depth map to enhance the training pipeline and 
JDACS~\cite{xu2021JDACS} proposes a unified framework to improve the robustness of self-supervisory signals against natural color disturbance in multi-view images.
U-MVS~\cite{xu2021u-mvs} utilizes the pseudo optical flow generated by off-the-shelf methods to improve the self-supervised model's performance. \cite{yang2021selfsup-cvp} renders pseudo depth labels from reconstructed mesh models and continues to train the self-supervised model.

\subsection{Knowledge Distillation}
Knowledge distillation~\cite{hinton2015distilling} aims to transfer knowledge from a teacher model to a student model, so that a powerful and lightweight student model can be obtained. \cite{park2019relational-KD,tung2019similarity-KD,peng2019correlation-KD,tian2019contrastive-KD,passalis2018learning-KD} consider knowledge at feature space and transfer it to the student model's feature space. Born-Again Networks (BAN)~\cite{furlanello2018born} trains a student model similarly parameterized as the teacher model and makes the trained student be a teacher model in a new round. The self-training scheme~\cite{xie2020self-train-distillation} generates distillation labels for unlabeled data and trains the student model with these labels.
Probabilistic knowledge transfer (PKT)~\cite{passalis2020probabilistic-KD,passalis2018pkt-KD} trains the student model via matching the probability distribution of the teacher model. Since labeled data are not required to minimize the difference of probability distribution, PKT can also be applied to unsupervised learning. In this work, we are inspired by PKT and offline distillation~\cite{romero2014fitnets-KD,zagoruyko2016paying-KD,huang2017like-KD,mirzadeh2020improved-KD,li2020few-KD} and propose to transfer the response-based knowledge~\cite{gou2021knowledge-survey} by forcing the predicted probability distribution of the student model to be similar to the probability distribution of the teacher model in an offline manner.


\begin{figure}[t]
\centering
\includegraphics[width=0.82\linewidth]{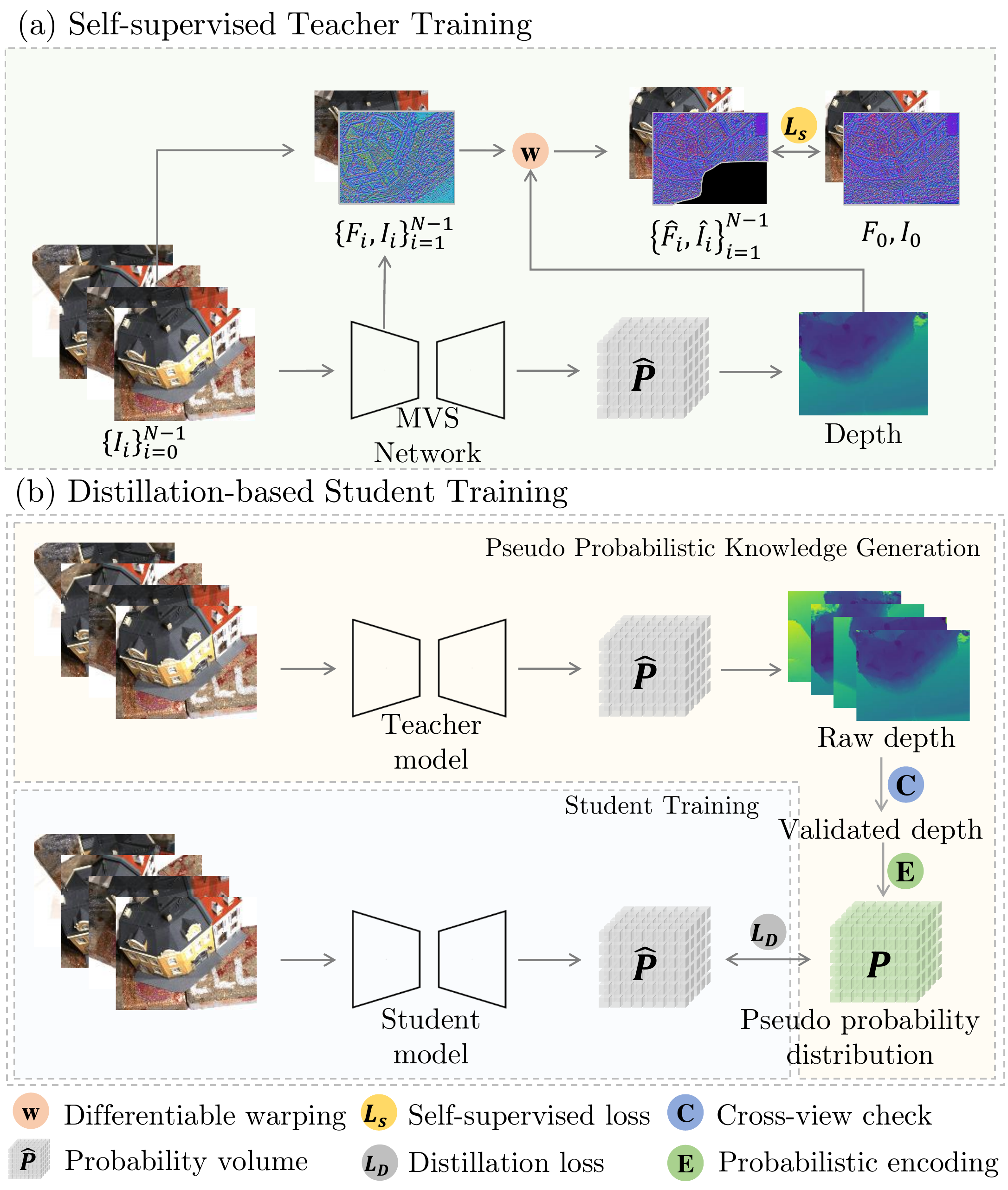}
\caption{Overview of KD-MVS. The first stage is self-supervised teacher training. The second stage is distillation-based student training, including pseudo probabilistic knowledge generation and student training.}
\label{fig:overview}
\end{figure}

\section{Methodology}
In this section, we elaborate the proposed training framework as illustrated in \cref{fig:overview}.
KD-MVS mainly consists of self-supervised teacher training (\cref{sec:unsup-training}) and distillation-based student training (\cref{sec:sd-training}).
Specifically, we first train a teacher model in a self-supervised manner by using both the photometric and featuremetric consistency between the reference view and the reconstructed views.
We then generate the pseudo probability distribution of the teacher model via cross-view check and probabilistic encoding. With the supervision of the pseudo probability, the student model is trained with distillation loss in an offline distillation manner.
It is worth noting that the proposed KD-MVS is a general pipeline for training MVS networks, it can be easily adapted to arbitrary learning-based MVS networks. In this paper, we mainly study KD-MVS with CasMVSNet~\cite{gu2020cascade}.


\subsection{Self-supervised Teacher Training}\label{sec:unsup-training}
In addition to conventional photometric consistency~\cite{khot2019Unsup-MVS} used in self-supervised MVS, we propose to use internal features and featuremetric consistency as an additional supervisory signal. Both the photometric and featuremetric consistency are obtained by calculating the distance between the reference view and the reconstructed views.
The following is the introduction to view reconstruction and loss formulation.

\subsubsection{View Reconstruction}

Given a reference image $\mathbf{I}_0$ and its neighboring source images $\{\mathbf{I}_i\}_{i=1}^{N-1}$, the common coarse-to-fine MVS network (e.g. CasMVSNet~\cite{gu2020cascade}) extracts features for all $N$ images at three different resolution levels ($1/4$, $1/2$, $1$), denoted as $\{\mathbf{F}_i^{1/4}, \mathbf{F}_i^{1/2}, \mathbf{F}_i\}_{i=0}^{N-1}$, and estimates the depth maps at these three corresponding levels, as $\mathbf{D}_0^{1/4}$, $\mathbf{D}_0^{1/2}$ and $\mathbf{D}_0$.

Taking $\mathbf{F}_0$ and $\mathbf{D}_0$ as an example, the warping between a pixel $\mathbf{p}$ at the reference view and its corresponding pixel $\hat{\mathbf{p}}_i$ at the $i$-th source view under estimated depth $d=\mathbf{D}_0(\mathbf{p})$ is defined as:
\begin{equation}
    \hat{\mathbf{p}}_{i} = \mathbf{K}_i[\mathbf{R_i}(\mathbf{K}_0^{-1}\mathbf{p}d)+\mathbf{t_i}],
    \label{Eq.warp}
\end{equation}
where $\mathbf{R}_i$ and $\mathbf{t}_i$ denote the relative rotation and translation from the reference view to the $i$-th source view. $\mathbf{K}_0$ and $\mathbf{K}_i$ are the intrinsic matrices of the reference and the $i$-th source camera. According to \cref{Eq.warp}, we are able to get the reconstructed images $\hat{\mathbf{I}}_i$ and features $\hat{\mathbf{I}}_i$ corresponding to the $i$-th source view. \cref{fig:warp} shows a photometric warping process from the $i$-th source view to the reference view.

\subsubsection{Loss Formulation}

\begin{figure}[t]
\centering
\includegraphics[width=\linewidth]{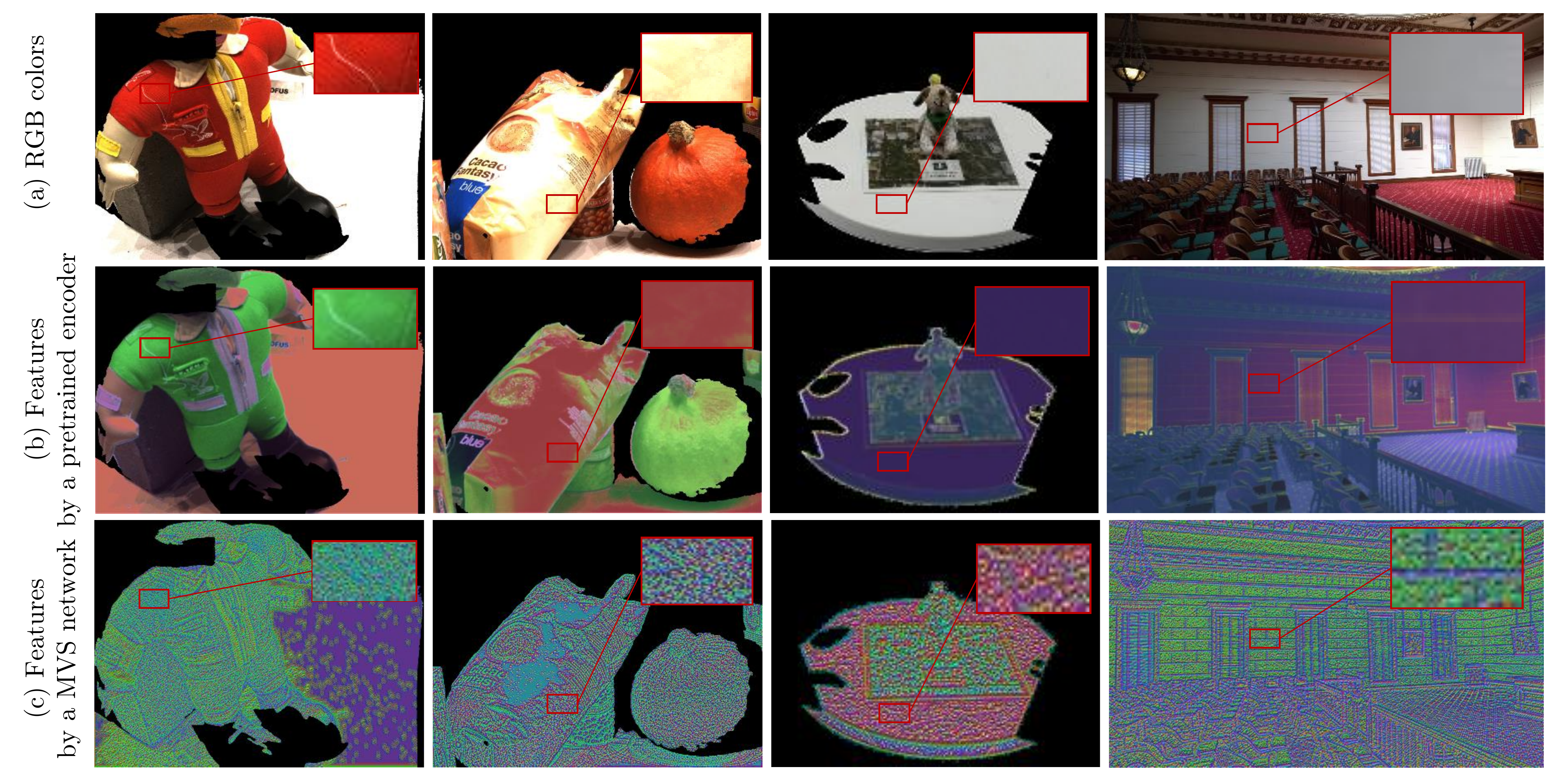}
\caption{Visualized examples of RGB colors (photometric) and extracted features (featuremetric). Dimension reduction of features is done by PCA. Features in (b) are extracted by a pre-trained ResNet-18~\cite{He2016resnet} and those in (c) are the features of the online training MVS network. Pretrained backbones tend to neglect the pixel-wise differences within intra-class regions while the online training MVS network is able to extract locally distinguishable features, which are more beneficial to downstream feature matching.
}
\label{fig:feature}
\end{figure}
Our self-supervised training loss consists of two components: photometric loss $\mathcal{L}_\textrm{photo}$ and featuremetric loss $\mathcal{L}_\textrm{fea}$. Following~\cite{khot2019Unsup-MVS}, the $\mathcal{L}_\textrm{photo}$ is based on the $\ell$-1 distance between the raw RGB reference image and the reconstructed images.
However, we find that the photometric loss is sensitive to lighting conditions and shooting angles, resulting in poor completeness of predictions.
To overcome this problem, we use the featuremetric loss to construct a more robust loss function.
Given the extracted features $\{\mathbf{F}_i\}_{i=0}^{N-1}$ from the feature net of MVS network, and the reconstructed feature maps $\hat{\mathbf{F}}_{i}$ generated from the $i$-th view, our featuremetric loss between $\hat{\mathbf{F}}_{i}$ and $\mathbf{F}_{0}$ is obtained by:
\begin{equation}
    \mathcal{L}_{\textrm{fea}}^{(i)} =  \|\mathbf{\hat{F}}_{i}-\mathbf{F}_0\|.
\end{equation}

It is worth noting that we put forward to use the internal features extracted by the internal feature net of the online training MVS network instead of the external features (e.g. extracted by a pre-trained backbone network~\cite{johnson2016perceptual-loss}) to compute featuremetric loss.
Our insight is that the nature of MVS is multi-view feature matching along epipolar lines, so the features are supposed to be locally discriminative. The pre-trained backbone networks, e.g. ResNet~\cite{He2016resnet} and VGG-Net~\cite{simonyan2014vgg}, are usually trained with image classification loss, so that their features are not locally discriminative.
As shown in \cref{fig:feature}, we compare the features extracted by an external pre-trained backbone (ResNet~\cite{He2016resnet}) and by the internal encoder of the MVS network during online self-supervised training. These two options lead to completely different feature representation and we study it in \cref{sec:ablation-feature} with experiments.

To summarize, the final loss function for self-supervised teacher training is
\begin{equation}
    \mathcal{L}_S = \frac{1}{|\mathbf{V}|} \sum_{\mathbf{p}\in\mathbf{V}} \sum_{i=1}^{N-1} (\lambda_{\textrm{fea}}\mathcal{L}_{\textrm{fea}}^{(i)}+\lambda_{\textrm{photo}}\mathcal{L}_{\textrm{photo}}^{(i)}),
\end{equation}
where $\mathbf{V}$ is the valid subset of image pixels. $\lambda_{\textrm{fea}}$ and $\lambda_{\textrm{photo}}$ are the two manually tuned weights, and in our experiments, we set them as 4 and 1 respectively. For coarse-to-fine networks, e.g. CasMVSNet, the loss function is applied to each of the regularization steps.

\begin{figure}[t]
\begin{minipage}[t]{0.49\linewidth}
    \centering
    \includegraphics[width=\linewidth]{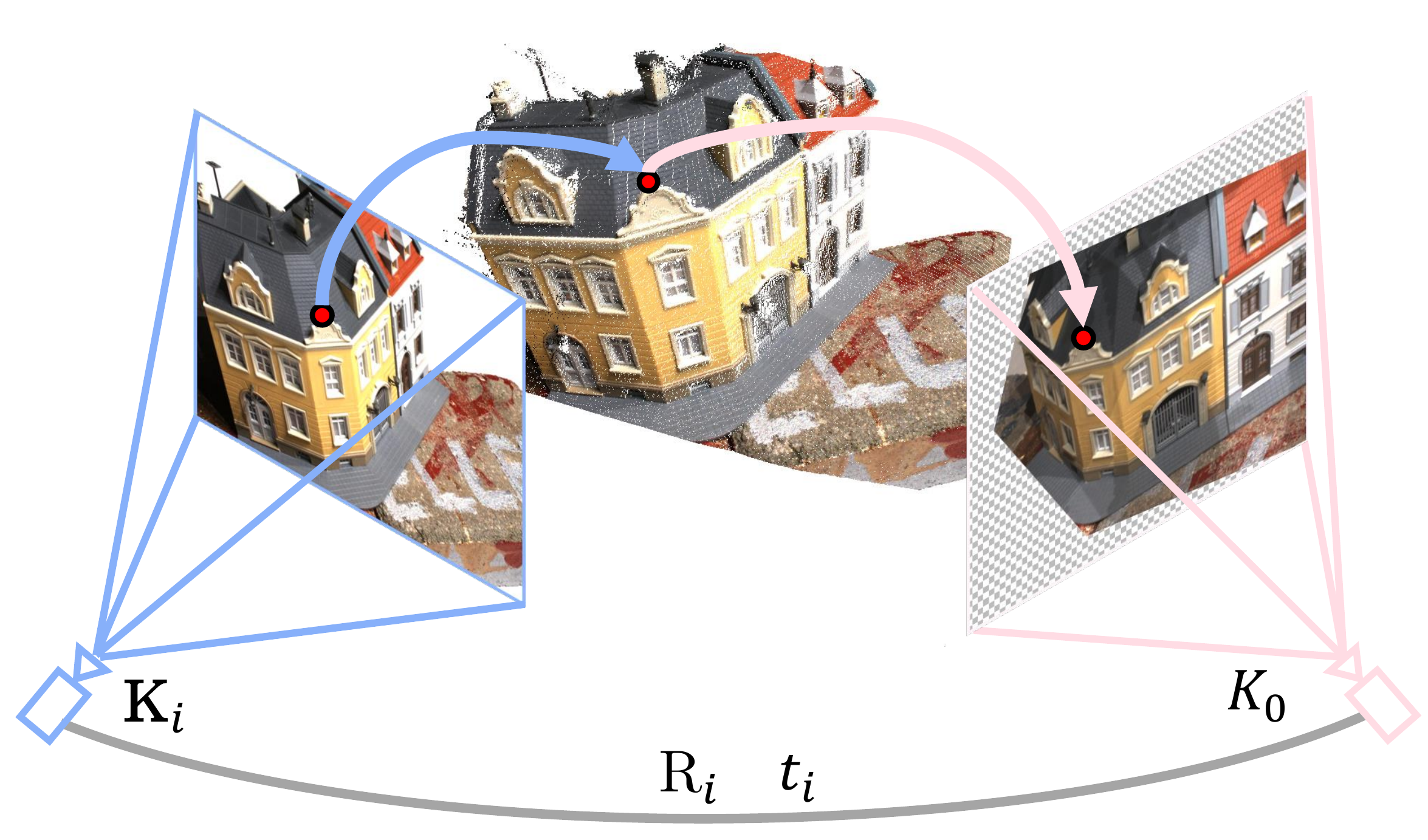}
    \caption{Photometric warping.}
    \label{fig:warp}
\end{minipage}\hfill
\begin{minipage}[t]{0.49\linewidth}
\centering
   \includegraphics[width=\linewidth]{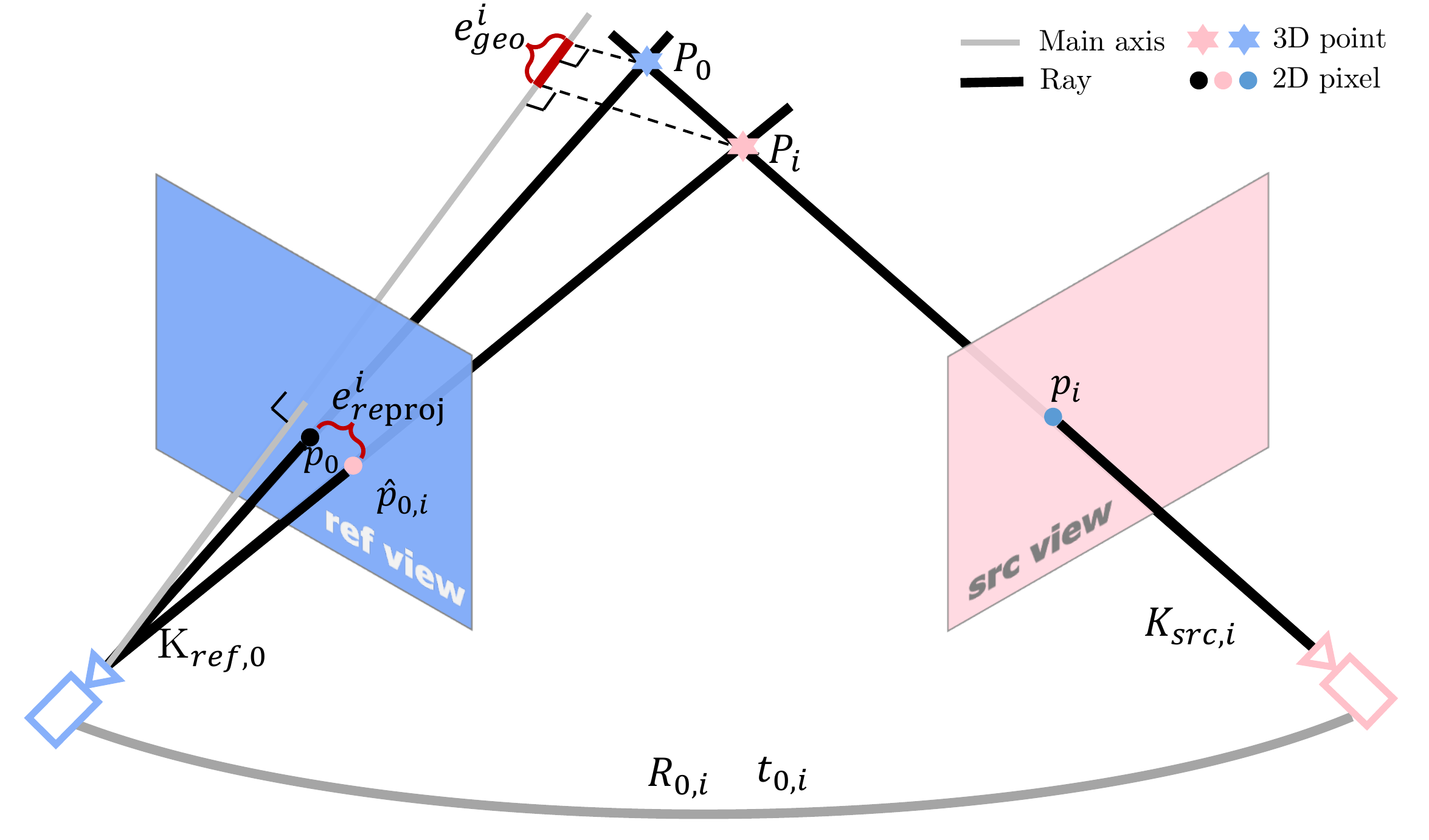}
    \caption{Cross-view check.}
    \label{fig:filter}
\end{minipage}
\end{figure}
\subsection{Distillation-based Student Training}\label{sec:sd-training}
To further stimulate the potential of the self-supervised MVS network, we adopt the idea of knowledge distillation and transfer the probabilistic knowledge of the teacher to a student model. This process mainly consists of two steps, namely pseudo probabilistic knowledge generation and student training.

\subsubsection{Pseudo Probabilistic Knowledge Generation}
We consider the knowledge transfer is done through the probability distribution as is done in~\cite{huang2017like-KD,romero2014fitnets-KD,tung2019similarity-KD}.
However, we face two main problems when applying distillation in MVS. (a) The raw per-view depth generated from the teacher model contains a lot of outliers, which is harmful to training student model. 
Thus we perform cross-view check to filter outliers. (b) The real probabilistic knowledge of the teacher model cannot be used directly to train the student model. 
That is because the depth hypotheses in the coarse-to-fine MVS network need to be dynamically sampled according to the results of the previous stage, and we cannot guarantee that the teacher model and student model always share the same depth hypotheses. 
To solve this problem, we propose to generate the pseudo probability distribution by probabilistic encoding.

\paragraph{Cross-view Check}\label{sec:cross-view-check}
is used to filter outliers in the raw depth maps, which are inferred by the self-supervised teacher model on the unlabeled training data.
Naturally, the outputs of the teacher model are per-view depth maps and the corresponding confidence maps. For coarse-to-fine methods, e.g. CasMVSNet~\cite{gu2020cascade}, we multiply confidence maps of all three stages to obtain the final confidence map and take the depth map in the finest resolution as the final depth prediction.

We denote the final confidence map of reference view as $\mathbf{C}_0$ and the final depth prediction as $\mathbf{D}_0$, the depth maps of source views as $\{\mathbf{D}_i\}_{i=1}^{N-1}$.
As is illustrated in \cref{fig:filter}, considering an arbitrary pixel $\mathbf{p}_0$ in the reference image coordinate, we cast the 2D point $\mathbf{p}_0$ to a 3D point $\mathbf{P}_0$ with the depth value $\mathbf{D}_0(\mathbf{p}_0)$. 
We then back-project $\mathbf{P}_0$  to $i$-th source view and obtain the point $\mathbf{p}_i$ in the source view. Using its estimated depth $\mathbf{D}_i(\mathbf{p}_i)$, we can cast the $\mathbf{p}_i$ to the 3D point $\mathbf{P}_i$. Finally, we back project $\mathbf{P}_i$ to the reference view and get $\hat{\mathbf{p}}_{0,i}$.
Then the reprojection error at $\mathbf{p}_0$ can be written as $e_{\textrm{reproj}}^{\textrm{i}}=\|\mathbf{p}_0-\hat{\mathbf{p}}_{0,i}\|$. A geometric error $e_{\textrm{geo}}^{\textrm{i}}$ is also defined to measure the relative depth error of $\mathbf{P}_0$ and $\mathbf{P}_i$ observed from the reference camera as $e_{\textrm{geo}}^{\textrm{i}}=\|\tilde{D}_0(\mathbf{P}_0)-\tilde{D}_0(\mathbf{P}_i)\|/\tilde{D}_0(\mathbf{P}_0)$, where the $\tilde{D}_0(\mathbf{P}_0)$ and $\tilde{D}_0(\mathbf{P}_i)$ are the projected depth of $\mathbf{P}_0$ and $\mathbf{P}_i$ in the reference view.
Accordingly, the validated subset of pixels with regard to the $i$-th source view is defined as
\begin{equation}
    \{\mathbf{p}_0\}_i=\{\mathbf{p}_0|\mathbf{C}_0(\mathbf{p}_0)>\tau_{\textrm{conf}},e_{\textrm{reproj}}^{\textrm{i}}<\tau_{\textrm{reproj}},e_{\textrm{geo}}^{\textrm{i}} <\tau_{\textrm{geo}}\},
\end{equation}
where $\tau$ represents threshold values, we set $\tau_{\textrm{conf}}$, $\tau_{\textrm{reproj}}$ and $\tau_{\textrm{geo}}$ to 0.15, 1.0 and 0.01 respectively. 
The final validated mask is the intersection of all $\{\mathbf{p}_0\}_i$ across $N$-1 source views. The obtained $\{\tilde{D}_0(\mathbf{P}_i)\}_{i=0}^{N-1}$ and validated mask will be further used to generate the pseudo probability distribution.

\paragraph{Probabilistic Encoding}
uses the $\{\tilde{D}_0(\mathbf{P}_i)\}_{i=0}^{N-1}$ to generate the pseudo probability distribution $P_{\mathbf{p}_0}(d)$ of depth value $d$ for each validated pixel $\mathbf{p}_0$ in reference view. 
We model $P_{\mathbf{p}_0}$ as a Gaussian distribution with a mean  depth value of $\mu({\mathbf{p}_0})$ and a variance of $\sigma^2({\mathbf{p}_0})$, which can be obtained by maximum likelihood estimation (MLE):
\begin{equation}
    \mu({\mathbf{p}_0}) = \frac{1}{N} \sum_{i=0}^{N-1}\tilde{D}_0(\mathbf{P}_i), \ \ \ 
    \sigma^2({\mathbf{p}_0}) = \frac{1}{N}\sum_{i=0}^{N-1}\left(\tilde{D}_0(\mathbf{P}_i) - \mu({\mathbf{p}_0})\right)^2.
\end{equation}
The $\mu({\mathbf{p}_0})$ fuses the depth information from multiple views, while the $\sigma^2({\mathbf{p}_0})$ reflects the uncertainty of the teacher model at $\mathbf{p}_0$, which will provide probabilistic knowledge for the student model during distillation training.

\subsubsection{Student Training}
With the pseudo probability distribution $P$, we are able to train a student model from scratch via forcing its predicted probability distribution $\hat{P}$ to be similar with $P$.
For the discrete depth hypotheses $\{d_k\}_{k=0}^D$, we obtain their pseudo probability $\{P(d_k)\}_{k=0}^D$ on the continuous probability distribution $P$ and normalize $\{P(d_k)\}_{k=0}^D$ using SoftMax, taking the result as the final discrete pseudo probability value.
We use Kullback–Leibler divergence to measure the distance between the student model's predicted probability and the pseudo probability.
The distillation loss $\mathcal{L}_{D}$ is defined as 
\begin{equation}
    \mathcal{L}_{D} = \mathcal{L}_{KL}(P||\hat{P}) = \sum_{\mathbf{p}\in \{\mathbf{p}_{v}\}} \left(P_\mathbf{p} - \hat{P}_\mathbf{p}\right)log\left(\frac{P_\mathbf{p}}{\hat{P}_\mathbf{p}}\right),
\end{equation}
where $\{\mathbf{p}_{v}\}$ represents the subset of valid pixels after cross-view check.

In experiments, we find that the trained student model also has the potential of becoming a teacher and further distilling its knowledge to another student model. As a trade-off between training time and performance, we perform the process of knowledge distillation once more. More details can be found in \cref{sec:iter}.


\setlength{\tabcolsep}{1pt}
\begin{table}[t]
\begin{center}
\caption{Quantitative results on DTU evaluation set~\cite{aanaes2016large} (\textbf{lower is better}). Sup. indicates whether the method is supervised or not.
}
\label{table:dtu-results}
\begin{tabular}{l|c|ccc}
\toprule
\textbf{Method} & Sup. & Acc. & Comp. & Overall\\
\hline
Gipuma~\cite{galliani2015massively}& - & 0.283 & 0.873 & 0.578  \\
COLMAP~\cite{schonberger2016mvs}& - & 0.400 & 0.664 & 0.532  \\
\hline
MVSNet~\cite{yao2018mvsnet} & \ding{51}  & 0.396 & 0.527 & 0.462 \\ 
AA-RMVSNet~\cite{wei2021aa}& \ding{51}   & 0.376 & 0.339 & 0.357  \\
CasMVSNet~\cite{gu2020cascade}& \ding{51}   & 0.325 &  0.385 & 0.355 \\ 
UCS-Net~\cite{cheng2020deep}& \ding{51}   & 0.338 & 0.349 & 0.344  \\
\hline
Unsup\_MVS~\cite{khot2019Unsup-MVS} & \ding{55} &  0.881 & 1.073 & 0.977 \\
MVS$^2$~\cite{dai2019mvs2} & \ding{55} &  0.760 & 0.515 & 0.637  \\
M$^3$VSNet~\cite{huang2020m3vsnet} & \ding{55} & 0.636 & 0.531 & 0.583  \\
JDACS~\cite{xu2021JDACS} & \ding{55} & 0.571 & 0.515 & 0.543  \\
{\scriptsize Self-supervised-CVP-MVSNet}~\cite{yang2021selfsup-cvp} & \ding{55} & \textbf{0.308} & 0.418 & 0.363 \\
U-MVS+MVSNet~\cite{xu2021u-mvs} & \ding{55} & 0.470 & 0.430 & 0.450 \\
U-MVS+CasMVSNet~\cite{xu2021u-mvs} & \ding{55} & 0.354 & 0.354 & 0.354 \\
\hline
\textbf{Ours+MVSNet} & \ding{55} & 0.424 & 0.426 & 0.425  \\
\textbf{Ours+CasMVSNet}& \ding{55}  &  0.359 & \textbf{0.295} & \textbf{0.327} \\
\bottomrule
\end{tabular}
\end{center}
\end{table}

\section{Experiments}

\subsection{Datasets}
DTU dataset~\cite{aanaes2016large} is captured under well-controlled laboratory conditions with a fixed camera rig, containing 128 scans with 49 views under 7 different lighting conditions. We split the dataset into 79 training scans, 18 validation scans, and 22 evaluation scans by following the practice of MVSNet~\cite{yao2018mvsnet}.
BlendedMVS dataset~\cite{yao2020blendedmvs} is a large-scale dataset for multi-view stereo and contains objects and scenes of varying complexity and scale. This dataset is split into 106 training scans and 7 validation scans.
Tanks and Temples benchmark~\cite{knapitsch2017tanks} is a public benchmark acquired in realistic conditions, which contains 8 scenes for the intermediate subset and 6 for the advanced subset.

\subsection{Implementation Details}
At the phase of self-supervised teacher training on DTU dataset~\cite{aanaes2016large}, we set the number of input images $N=5$ and image resolution as $512\times 640$. For coarse-to-fine regularization of CasMVSNet~\cite{gu2020cascade}, the settings of depth range and the number of depth hypotheses are consistent with \cite{gu2020cascade}; the depth interval decays by 0.25 and 0.5 from the coarsest stage to the finest stage.
The teacher model is trained with Adam for 5 epochs with a learning rate of 0.001.
At the phase of distillation-based student training, we train the student model with the pseudo probability distribution for 10 epochs. Model training of all experiments is carried out on 8 NVIDIA RTX 2080 GPUs.

\subsection{Experimental Results}

\begin{figure}[t]
\centering
\includegraphics[width=0.9\linewidth]{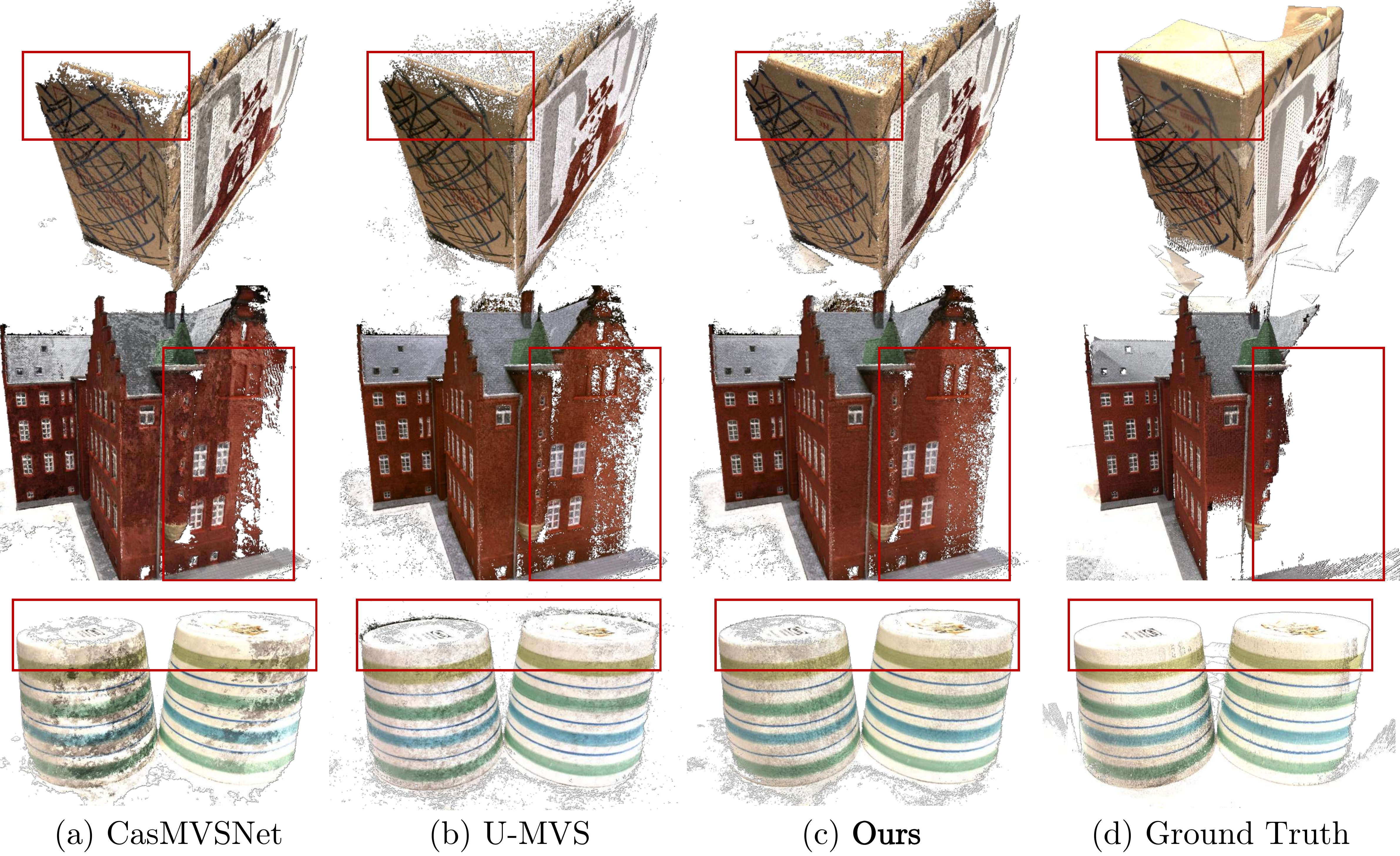}
\caption{Comparison of reconstructed results with the supervised baseline~\cite{gu2020cascade} and the state-of-the-art self-supervised method U-MVS~\cite{xu2021u-mvs} on DTU test set~\cite{aanaes2016large}.}
\label{fig:pcd-dtu}
\end{figure}

\subsubsection{DTU Dataset}
We evaluate KD-MVS, applied to MVSNet~\cite{yao2018mvsnet} and CasMVSNet~\cite{gu2020cascade} on DTU dataset~\cite{aanaes2016large}. 
We set $N = 5$ and input resolution as $864 \times 1152$ at evaluation. 
Quantitative comparisons are shown in \cref{table:dtu-results}. Accuracy, Completeness and Overall are the three official metrics from~\cite{aanaes2016large}.
Our method outperforms all self-supervised methods by a large margin and even the supervised ones.
\cref{fig:pcd-dtu} shows a visualization comparison of reconstructed point clouds. Our method achieves much better reconstruction quality when compared with the baseline network and the state-of-the-art self-supervised method.

\begin{table}[t]
\centering
\caption{Quantitative results on the intermediate set of Tanks and Temples benchmark~\cite{tnt}. Sup. indicates whether the method is supervised or not. \textbf{Bold} and \underline{underlined} figures indicate the best and the second best results.}
\label{table:tnt-inter}
\resizebox{\linewidth}{!}{
\begin{tabular}{l|c|c|cccccccc}
\toprule
\textbf{Method}& Sup. &  Mean & Family & Francis & Horse & L.H. & M60 & Panther & P.G. & Train \\
\hline
COLMAP~\cite{schonberger2016mvs}& -  & 42.14 & 50.41 & 22.25 & 26.63 & 56.43 & 44.83 & 46.97 & 48.53 & 42.04\\
ACMM~\cite{xu2019multi}& - &57.27	&	69.24	&51.45	&46.97	&63.20&	55.07&	57.64&	60.08&	54.48 \\
AttMVS~\cite{luo2020attention}& -  &	{60.05} &	73.90 &	 62.58 &	44.08 &	\textbf{64.88} &	56.08&	59.39 &	 \textbf{63.42} & \underline{56.06}\\
\hline
MVS$^2$~\cite{dai2019mvs2} & \ding{55} & 37.21 & 47.74 & 21.55 & 19.50 & 44.54 & 44.86 & 46.32 & 43.38 & 29.72\\
M$^3$VSNet~\cite{huang2020m3vsnet} & \ding{55} & 37.67 & 47.74 & 24.38 & 18.74 & 44.42 & 43.45 & 44.95 & 47.39 & 30.31 \\
JDACS~\cite{xu2021JDACS} & \ding{55} & 45.48 & 66.62 & 38.25 & 36.11 & 46.12 & 46.66 & 45.25 & 47.69 & 37.16 \\
{\scriptsize Self-supervised-CVP-MVSNet}~\cite{yang2021selfsup-cvp} & \ding{55} &  46.71& 64.95& 38.79&   24.98& 49.73& 52.57&51.53&50.66&40.52 \\
U-MVS+CasMVSNet~\cite{xu2021u-mvs} & \ding{55} & 57.15 & 76.49 & 60.04 & 49.20 & 55.52 & 55.33 & 51.22 & 56.77 & 52.63 \\
\hline
CasMVSNet~\cite{gu2020cascade}& \ding{51}  & 56.84	&	76.37&	58.45&	46.26&	55.81&	56.11&	54.06&	58.18&	49.51 \\
Vis-MVSNet~\cite{zhang2020visibility}& \ding{51} & 60.03 &	77.40&	60.23&	47.07&	63.44&	62.21&	57.28&	{60.54}&	52.07\\
AA-RMVSNet~\cite{wei2021aa}& \ding{51}  & 61.51 & 77.77 &	59.53&	51.53 &	64.02&	\underline{64.05} & 59.47&	60.85&	54.90\\
EPP-MVSNet~\cite{ma2021epp}& \ding{51}  & \underline{61.68} & \underline{77.86} & \underline{60.54} & \underline{52.96} & 62.33 & 61.69 & \underline{60.34} & \underline{62.44} & 55.30 \\
\hline
\textbf{Ours+CasMVSNet} & \ding{55} & \textbf{64.14} & \textbf{80.42} & \textbf{67.42} & \textbf{54.02} & \underline{64.52} & \textbf{64.18} & \textbf{61.60} & 62.37 & \textbf{58.59}\\
\bottomrule
\end{tabular}}
\end{table}

\begin{table}[t]
\centering
\caption{Quantitative results on the advanced set of Tanks and Temples benchmark~\cite{tnt}.}
\label{table:tnt-adv}
\resizebox{\linewidth}{!}{
\begin{tabular}{l|c|c|cccccc}
\toprule
\textbf{Method}& Sup. &  Mean & Auditorium & Ballroom & Courtroom & Museum & Palace & Temple \\
\hline 
COLMAP~\cite{schonberger2016mvs}& -  & 27.24 & 16.02 & 25.23 & 34.70 & 41.51 & 18.05 & 27.94\\
ACMM~\cite{xu2019multi}& - & 34.02 & 23.41&32.91& \textbf{41.17} &48.13&23.87&34.60 \\
\hline
CasMVSNet~\cite{gu2020cascade}& \ding{51}  & 31.12 & 19.81 & 38.46 & 29.10 & 43.87 & 27.36 & 28.11 \\
AA-RMVSNet~\cite{wei2021aa}& \ding{51}  & 33.53&20.96 & \underline{40.15} & 32.05 & 46.01 & 29.28 & 32.71\\
Vis-MVSNet~\cite{zhang2020visibility}& \ding{51} & 33.78 & 20.79 &	38.77 &	32.45 & 44.20 & 28.73 & \underline{37.70}\\
EPP-MVSNet~\cite{ma2021epp}& \ding{51}  & 35.72 & \underline{21.28} & 39.74 & 35.34 & \textbf{49.21} & \underline{30.00} & \textbf{38.75} \\
\hline
\textbf{Ours+CasMVSNet} & \ding{55} & \textbf{37.96} & \textbf{27.22} & \textbf{44.10} & \underline{35.53} & \underline{49.16} & \textbf{34.67} & 37.11\\
\bottomrule
\end{tabular}}
\end{table}

\subsubsection{Tanks and Temples Benchmark}
\begin{figure}[t]
\centering
\includegraphics[width=0.95\linewidth]{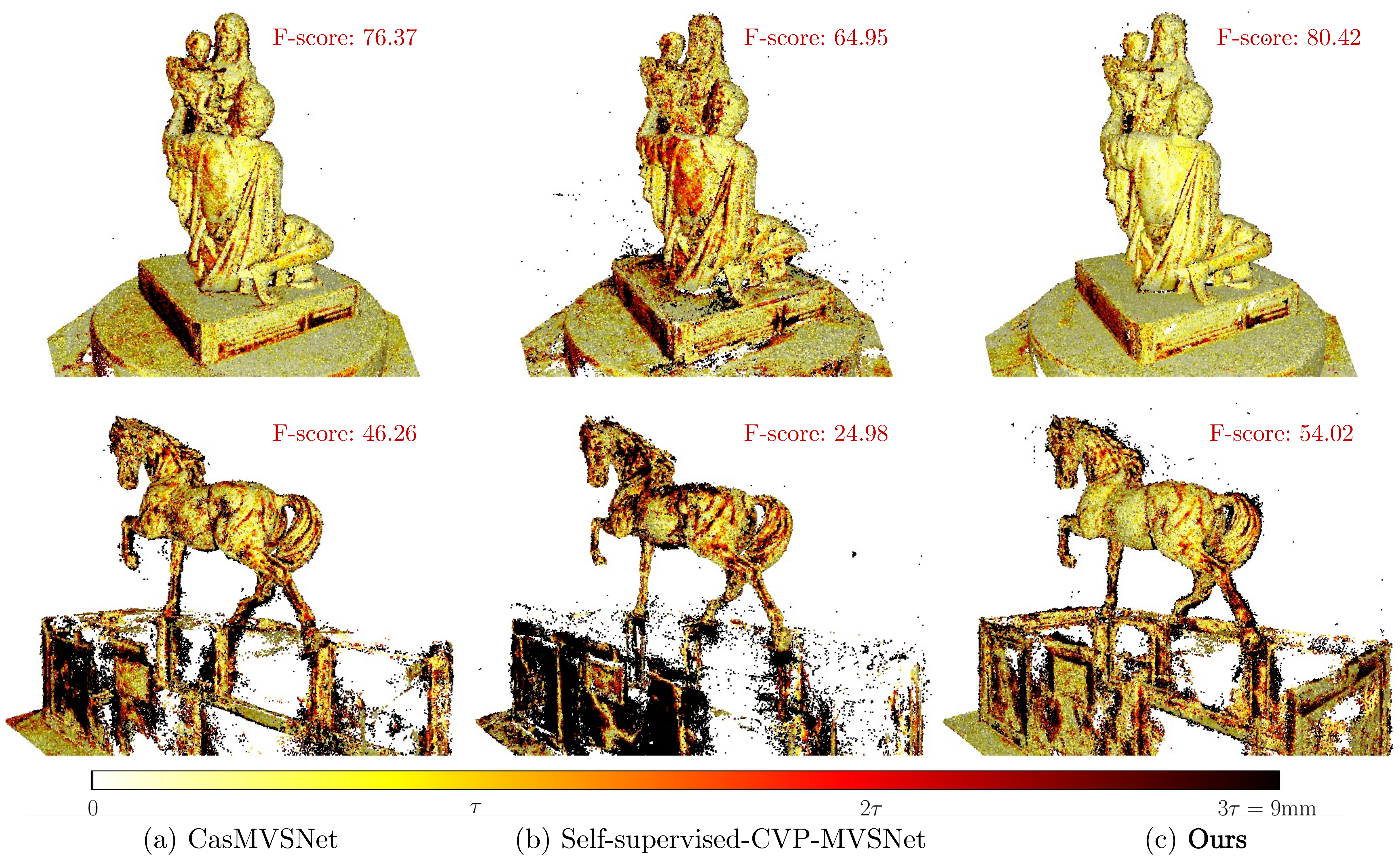}
\caption{Comparison of reconstructed results with the supervised baseline CasMVSNet~\cite{gu2020cascade} and the state-of-the-art self-supervised method~\cite{yang2021selfsup-cvp} on Tanks and Temples benchmark~\cite{knapitsch2017tanks}. $\tau=3mm$ is the distance threshold determined officially and darker regions indicate larger error encountered with regard to $\tau$.
}
\label{fig:pcd-tnt}
\end{figure}

We test our method on Tanks and Temples benchmark~\cite{knapitsch2017tanks} to demonstrate the ability to generalize on varying data. 
For a fair comparison with state-of-the-art methods, we fine-tune our model on the training set of the BlendedMVS dataset~\cite{yao2020blendedmvs} using the original image resolution ($576\times 768$) and $N = 5$.
More details about the fine-tuning process can be found in supp. materials.
Similar to other methods~\cite{gu2020cascade,xu2021u-mvs}, the camera parameters, depth ranges, and neighboring view selection are aligned with \cite{yao2019recurrent}. We use images of the original resolution for inference. Quantitative results are shown in \cref{table:tnt-inter} and \cref{table:tnt-adv}, and the qualitative comparisons ares shown in \cref{fig:pcd-tnt}.

\subsubsection{BlendedMVS Dataset}
We further demonstrate the quality of depth maps on the validation set of BlendedMVS dataset~\cite{yao2020blendedmvs}. The details of the training process can be found in supp. materials.
We set $N=5$, image resolution as $512\times 640$, and apply the evaluation metrics described in \cite{darmon2021deep} where depth values are normalized to make depth maps with different depth ranges comparable.
Quantitative results are illustrated in \cref{table:bld-results}. EPE stands for the endpoint error, which is the average $\ell$-1 distance between the prediction and the ground truth depth; $e_1$ and $e_3$ represent the percentage of pixels with depth error larger than 1 and larger than 3.

\begin{table}[t]
\begin{minipage}[t]{0.49\linewidth}
\centering
\caption{Quantitative results towards predicted depth maps on BlendedMVS validation set~\cite{yao2020blendedmvs} (\textbf{lower is better}).}
\label{table:bld-results}
\begin{tabular}{l|c|ccc}
\toprule
\textbf{Method} & Sup. & EPE & $e_1$ & $e_3$ \\
\hline
MVSNet~\cite{yao2018mvsnet}& \ding{51} & 1.49 & 21.98 & 8.32 \\
CVP-MVSNet~\cite{yang2020cost}& \ding{51} & 1.90 &  19.73 & 10.24 \\
CasMVSNet~\cite{gu2020cascade}& \ding{51} & 1.43 &  19.01 & 9.77 \\ 
Vis-MVSNet~\cite{zhang2020visibility}& \ding{51} & 1.47  & 15.14 & 5.13 \\
EPP-MVSNet~\cite{ma2021epp}& \ding{51} & 1.17 & 12.66 & 6.20 \\
\hline
\textbf{Ours}& \ding{55} & \textbf{1.04}  & \textbf{10.17} & \textbf{4.94} \\
\bottomrule
\end{tabular}
\end{minipage}\hfill
\begin{minipage}[t]{0.49\linewidth}
    \centering
    \caption{Ablation study on loss for self-supervised training stage (teacher model). $\mathcal{L}_{\textrm{fea}}$ and $\mathcal{L}_{\textrm{fea}}^*$ denotes featuremetric loss by the internal feature encoder and by an external pretrained encoder (ResNet-18~\cite{He2016resnet}) respectively.}
    \label{table:feature}
    \begin{tabular}{ccc|ccc}
    \toprule
    $\mathcal{L}_{\textrm{photo}}$ & $\mathcal{L}_{\textrm{fea}}^*$ & $\mathcal{L}_{\textrm{fea}}$ & Acc. & Comp. & Overall \\
    \hline
    \ding{51} & & & 0.489 & 0.501 & 0.495\\
    \ding{51}& \ding{51} & & 0.477 & 0.441 & 0.459  \\
    \ding{51}&  & \ding{51} & \textbf{0.457} & \textbf{0.399} & \textbf{0.428}\\
    \bottomrule
    \end{tabular}
\end{minipage}
\end{table}

\begin{table}[t]
\begin{minipage}[t]{0.49\linewidth}
\centering
\label{table:ablation-iter}
\caption{Ablation study on the number of iterations for distillation training. Note that we consider the number of distillation rounds equal to the number of times fused depth is generated and verified.}
\begin{tabular}{c|ccc}
\toprule
\#round(s)  &  Acc. & Comp. & Overall\\
\hline
1 & 0.387 & 0.334 & 0.361 \\
2 & 0.359 & \textbf{0.295} & \textbf{0.327}\\
3 & \textbf{0.357} & 0.298 & \textbf{0.327}\\
4 & \textbf0.358 & 0.297 & 0.328\\
\bottomrule
\end{tabular}
\end{minipage}\hfill
\begin{minipage}[t]{0.49\linewidth}
\centering
\caption{Ablation study on the main factor of effectiveness. Mask indicates whether to use the validated mask. Depth indicates using ground truth depth or validated depth. Loss indicates which loss is used.}
\label{table:dark-knowledge}
\begin{tabular}{l|ccc|ccc}
\toprule
 & Mask & Depth & Loss & Acc. & Comp. & Over. \\
\hline
(1) & \ding{55}& GT & $\ell$-1 & 0.358  & 0.346  & 0.352 \\
(2) & \ding{51}& GT & $\ell$-1 &  \textbf{0.352} & 0.334  & 0.343 \\
(3) & \ding{51}& vali. & $\ell$-1  & 0.361 & 0.331  & 0.346 \\
(4) & \ding{51}& vali. & distill & 0.359  & \textbf{0.295} & \textbf{0.327} \\
\bottomrule
\end{tabular}
\end{minipage}
\end{table}

\subsection{Ablation Study}

\subsubsection{Implementation of Featuremetric Loss}\label{sec:ablation-feature}
As analyzed in \cref{sec:unsup-training}, we consider the nature of the MVS is multi-view feature matching along epipolar lines, where the features are supposed to be relatively locally discriminative.
\cref{table:feature} shows the quantitative results of different settings.
Compared with using photometric loss only, both internal featuremetric and external featuremetric loss can boost the performance. And our proposed internal featuremetric loss shows superiority over the external featuremetric loss with external features by a pre-trained ResNet.
It is worth noting that it is not feasible to adopt our featuremetric loss alone. The reason is that the feature network is online trained within the MVS network and thus applying featuremetric loss alone will lead to failure of training where features tend to be a constant (typically 0).

\subsubsection{Number of Self-training Iterations}
\label{sec:iter}
Given the scheme of knowledge distillation via generating pseudo probability, we can iterate the distillation-based student training for an arbitrary number of loops. Here we study the performance gain when the number of iterations increases in Tab. \textcolor{red}{6}. As a trade-off of efficiency and accuracy, we set the number of iterations to be 2.


\section{Discussion}
\subsection{Insights of Effectiveness}
\begin{figure}[t]
\centering
\includegraphics[width=1.0\linewidth]{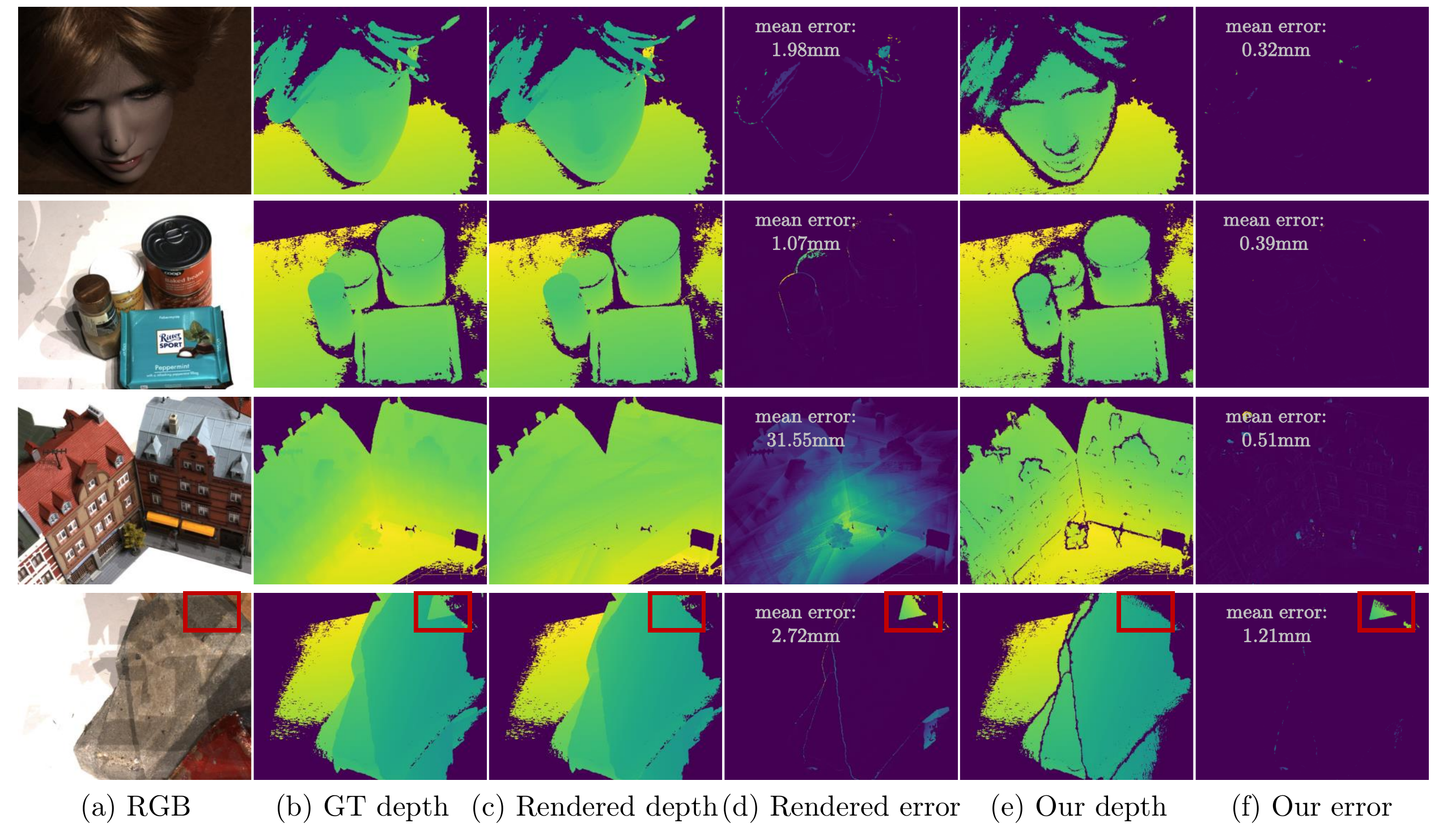}
\caption{Visualization of depth maps and errors. (a) RGB reference images; (b) ground-truth depth maps; (c) rendered depth maps by \cite{yang2021selfsup-cvp}; (d) errors between (b) and (c); (e) pseudo labels in KD-MVS; (f) errors between (b) and (e). We apply the same mask on (b)(c)(e) for better visualization.
}
\label{fig:depth}
\end{figure}

We attribute the effectiveness of KD-MVS to the following four parts. (a) The first one is multi-view consistency as introduced in \cref{sec:cross-view-check}, which can be used to filter the outliers in noisy raw depth maps. The remaining inliers are relatively accurate and are equivalent to ground-truth depth to a certain extent.
(b) The probabilistic knowledge brings performance gain to the student model.
Compared with using hard labels such as $\ell$-1 loss and depth labels, applying soft probability distribution to student model brings additional inter-depth relationships and thus reduces the ambiguity of noisy 3D points.
(c) The validated depth contains less perspective error than rendered ground truth labels. As shown in the last row of \cref{fig:depth} (marked with a red box), there are some incorrect values in the ground-truth depth maps of DTU dataset~\cite{aanaes2016large} caused by perspective error, which is harmful to training MVS models. 
(d) The validated masks of the teacher model reduce the ambiguity of prediction by filtering the samples which are hard to learn, benefiting the convergence of the model.
We perform an ablation study on these parts as shown in \cref{table:dark-knowledge}. (1) and (2) show that the validated mask is helpful, (3) and (4) show that enforcing the probability distribution can bring significant improvement. More details can be found in supp. materials.

\subsection{Comparisons to SOTA Methods}
\subsubsection{U-MVS}\cite{xu2021u-mvs} leverages optical flow to compute a depth-flow consistency loss. To get reliable optical flow labels, U-MVS trains a PWC-Net~\cite{sun2018pwc} on DTU dataset~\cite{aanaes2016large} in a self-supervised manner, which costs additional training time and needs storage space for the pseudo optical flow labels (more than 120$GB$).

\subsubsection{Self-supervised-CVP-MVSNet}\cite{yang2021selfsup-cvp}
renders depth maps from the reconstructed meshes, which brings in error during Poisson reconstruction~\cite{2013screened}. We compare the rendered depth maps \cite{yang2021selfsup-cvp} and our validated depth maps in \cref{fig:depth}.

\subsection{Limitations}
\begin{itemize}
    \item[-] The quality of pseudo probability distribution highly depends on the cross-view check stage and relevant hyperparameters need to be tuned carefully.
    \item[-] Knowledge distillation is known as data-hungry and it may not work as expected with a relatively small-scale dataset.
\end{itemize}


\section{Conclusion}
In this paper, we propose KD-MVS, which is a general self-supervised pipeline for MVS networks without any ground-truth depth as supervision. In the self-supervised teacher training stage, we leverage a featuremetric loss term, which is more robust than photometric loss alone. The features are yielded internally by the MVS network itself, which is end-to-end trained under implicit supervision. To explore the potential of self-supervised MVS, we adopt the idea of knowledge distillation and distills the teacher's knowledge to a student model by generating pseudo probability distribution.
Experimental results indicate that the self-supervised training pipeline has the potential to obtain reconstruction quality equivalent to supervised ones.

%
%
\bibliographystyle{splncs04}
\bibliography{egbib}


\clearpage
\section*{Appendix}
\appendix
\section{More Insights of KD-MVS}\label{sec:1.1}

In this section, we discuss the potential reasons why self-supervised methods can obtain comparable (even better) results compared to supervised methods. Here we elaborate this from the following two perspectives.

(a) Self-supervised methods are able to generate accurate pseudo labels with cross-view check. According to Eq. (4) of the paper, only the inliers (whose depth prediction is accurate) can be kept given a strict threshold. 
We also visualize the depth error of the pseudo depth in Fig. 8, which is relevant to the overall error of point cloud results. The first and the second row indicate the pseudo depth has already been to some extent accurate in most scenes. Similar conclusion can also be inferred from the Tab. 7(2)(3) of the paper.

\begin{figure}[h]
\centering
\includegraphics[width=0.99\linewidth]{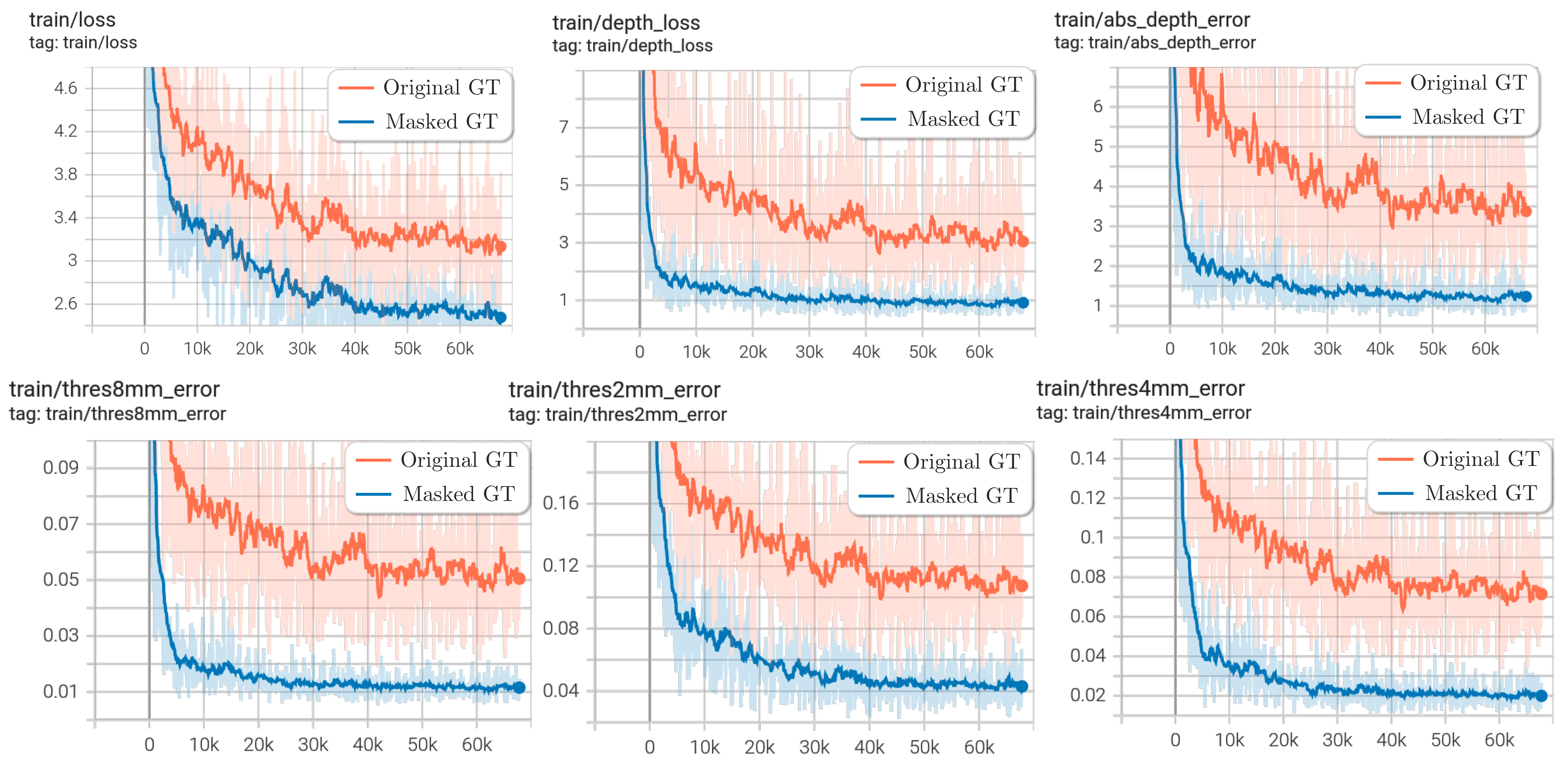}
\caption{Comparisons of training processes in several metrics. \textcolor[RGB]{255,112,76}{Original GT} indicates training with the original ground-truth depth of DTU dataset~\cite{aanaes2016large}. \textcolor[RGB]{1,118,184}{Masked GT} is obtained by masking the original GT with the mask of pseudo depth. Reducing the toxic samples makes the training phase more stable.}
\label{fig:figure_loss}
\vspace{-2pt}
\end{figure}

\begin{figure}[t]
\centering
\includegraphics[width=0.65\linewidth]{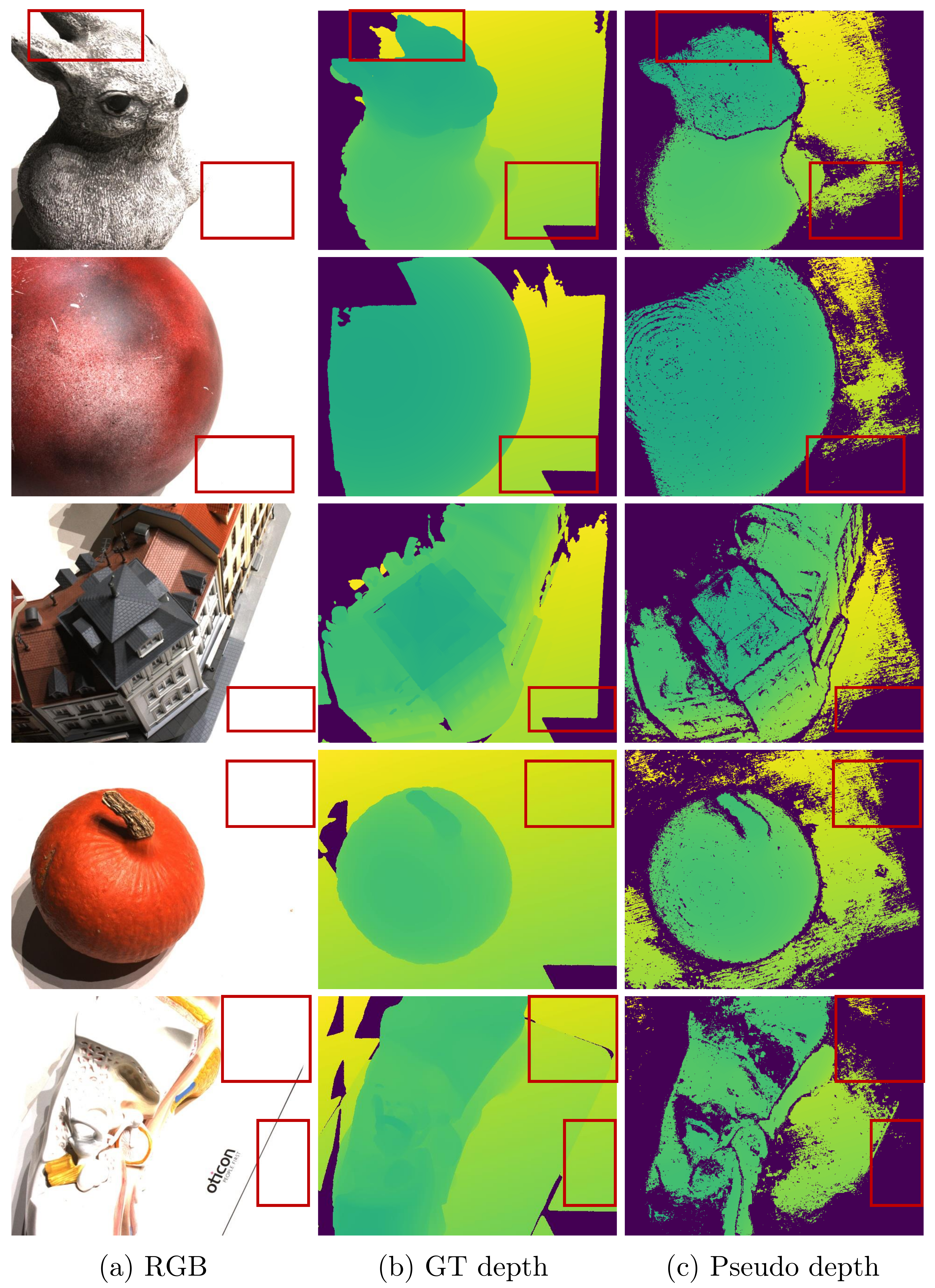}
\caption{Visualized comparisons between GT depth and pseudo depth. Red boxes indicate textureless regions, which are nearly unpredictable. Pseudo depth maps contain less textureless (misleading) regions by filtering the outliers with cross-view check.}
\label{fig:depth_supp}
\vspace{-8pt}
\end{figure}

(b) The pseudo labels have advantages over GT in some aspects.
Generally speaking, the datasets contain many pixels (namely training samples) that are normally textureless and considered as ``toxic" for training.  For example, if we force the network to estimate depth values for a purely white region, which should be inherently unpredictable, the network will be confused since there are simply no valid features extracted. \cref{fig:depth_supp} shows visualized comparisons of GT depth and pseudo depth, the red boxes highlight the textureless regions, which are nearly unpredictable. The pseudo depth maps contain fewer misleading regions by filtering the outliers with cross-view check.
By reducing the toxic samples, the training process will be more stable and the performance will be improved.
\cref{fig:figure_loss} shows the comparison of several metrics in the training phase. Compared with using the original depth (the orange curve), reducing the toxic samples by using the mask of the pseudo depth makes the training phase more stable and helps the student model converge faster.

It is worth noting that similar results can be found in~\cite{xu2021u-mvs} and~\cite{yang2021selfsup-cvp}, which also use the pseudo depth labels to train MVS network in a self-supervised manner and achieve better performance than the supervised baseline methods. 

Besides, we propose to leverage the probabilistic knowledge, which is verified to be effective in classification task~\cite{li2020few-KD,huang2017like-KD,passalis2018pkt-KD,tung2019similarity-KD}. Some works~\cite{hinton2015distilling,xie2020self-train-distillation} call this probabilistic knowledge as dark knowledge and believe they contain inter-class information. As MVS can also be handled as a classification task, it can benefit from probabilistic knowledge too.

\section{More Details of Experiment Settings}
\subsection{Fine-tuning on BlendedMVS}\label{sec:2.1}
As done in many supervised methods~\cite{wei2021aa,zhang2020visibility,ding2021transmvsnet}, we train the student model on BlendedMVS dataset~\cite{yao2020blendedmvs} for better performance and fair comparison. Concretely, we use the student model trained on DTU dataset~\cite{aanaes2016large} as the teacher model and generate pseudo probabilistic labels for BlendedMVS dataset. A new student model is then trained using the pseudo labels of both BlendedMVS and DTU from scratch. The process of fine-tuning still follows the self-supervised scheme of KD-MVS and leverages no extra manual annotation.

\subsection{Ablation Study on the Insights of effectiveness}\label{sec:2.2}
We perform ablation study to verify the insights of effectiveness in Sec. 5.1 of the paper, and the results are listed in Tab. 7.  The $Mask$ indicates whether use the validated masks, which are generated by performing cross-view check on the raw depth maps of the teacher model. For (2) in Tab. 7, since the GT depth itself has a mask, we take the intersection of the two masks as the final mask. The $Depth$ indicates which kind of depth is used to train the student model. Comparing the results of (2) and (3), it can be found that the pseudo depth label is accurate. The $Loss$ indicates which kind of loss function is used to train the student model. When the GT depth is used, we can only use the $\ell$-1 loss. When the distillation loss is used, we use the probabilistic encoding to generate the pseudo probability distribution. Comparing the results of (3) and (4), we can find that the distillation loss brings a big performance gain.

\section{Ablation Study on Hyper-parameters}

\subsection{Number of Views \& Input Resolution}
We perform ablation study against the number of input views $N$ and input resolution $H\times W$ on DTU evaluation set~\cite{aanaes2016large}, and the results are shown in \cref{tab:NHW}.

\begin{table}[t]
    \footnotesize
    \setlength\tabcolsep{3pt}
    \centering
    \caption{Ablation study on number of input views $N$ and image resolution $H\times W$ on DTU evaluation set~\cite{aanaes2016large} (\textbf{lower is better}). We use the same student model and keep the other settings fixed.}
    \label{tab:NHW}
    \begin{tabular}{c|c|ccc}
    \toprule
    $N$ & $H\times W$  & Acc.(mm) & Comp.(mm) & Overall(mm) \\
    \hline
    3 &  $864\times 1152$  & 0.366 & 0.316 & 0.341  \\
    5 &  $864\times 1152$  & \textbf{0.359} & \textbf{0.295} & \textbf{0.327}  \\
    7 &  $864\times 1152$  & \textbf{0.359} & 0.299 & 0.329 \\
    9 &  $864\times 1152$  & 0.360 & 0.301 & 0.331  \\
    5 &  $512\times 640$   & 0.405 & 0.313 & 0.359  \\
    \toprule
    \end{tabular}
    \vspace{-8pt}
\end{table}

\subsection{Thresholds in Cross-view Check}
As introduced in Sec. 5.3 of the paper and~\cref{sec:1.1} of the supp. materials, the quality of pseudo probability distribution depends on the cross-view check strategy and the relevant hyper-parameters. We perform ablation study on the three threshold parameters in cross-view check, and the results are shown in~\cref{tab:thershold}.
\begin{table}[t]
    \footnotesize
    \setlength\tabcolsep{2pt}
    \centering
    \caption{Ablation study on different settings of the threshold parameters in cross-view check. We use the same raw depth of the self-supervised teacher model to genrate the pseudo labels, and show the results of the student model on DTU evaluation set~\cite{aanaes2016large} (\textbf{lower is better}).}
    \label{tab:thershold}
    \begin{tabular}{c|ccc|ccc}
    \toprule
     & $\tau_{\textrm{conf}}$ & $\tau_{\textrm{reproj}}$ & $\tau_{\textrm{geo}}$ & Acc.(mm) & Comp.(mm) & Overall(mm) \\
    \hline
    (a) & 0.30 & 2.0 & 0.020 & 0.410 & 0.382 & 0.396 \\
    (b) & 0.20 & 1.5 & 0.015 & 0.384 & 0.320 & 0.352 \\
    (c) & 0.15 & 1.0 & 0.010 & 0.359 & \textbf{0.295} & \textbf{0.327} \\
    (d) & 0.10 & 0.5 & 0.005 & \textbf{0.350} & 0.306 & 0.328 \\
    \toprule
    \end{tabular}
    \vspace{-8pt}
\end{table}

\section{More Point Cloud Results}
\begin{figure}[t]
\centering
\includegraphics[width=1.0\linewidth]{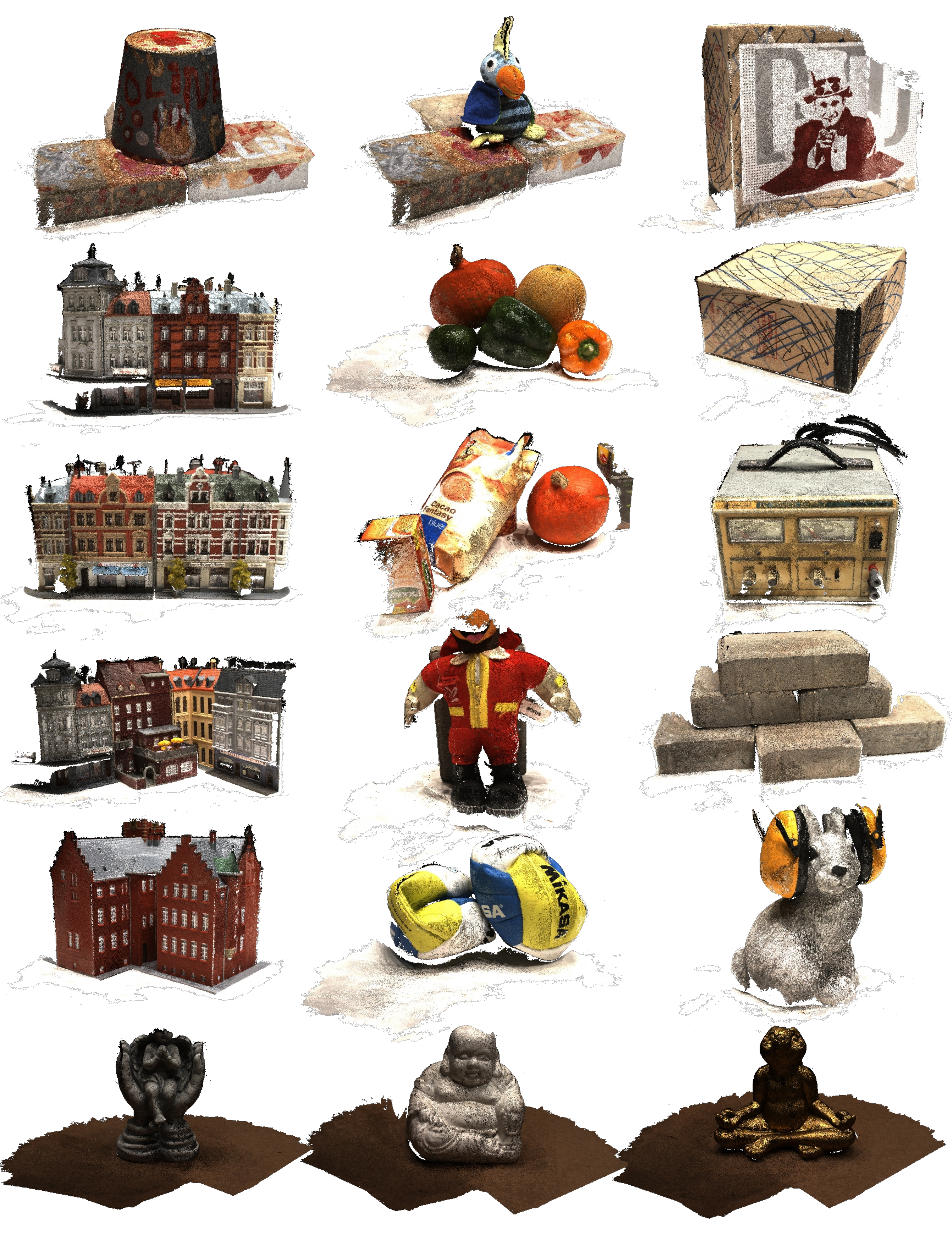}
\caption{Point cloud results on DTU dataset~\cite{aanaes2016large}.}
\label{fig:dtu-pcd}
\vspace{-8pt}
\end{figure}

\begin{figure}[t]
\centering
\includegraphics[width=0.95\linewidth]{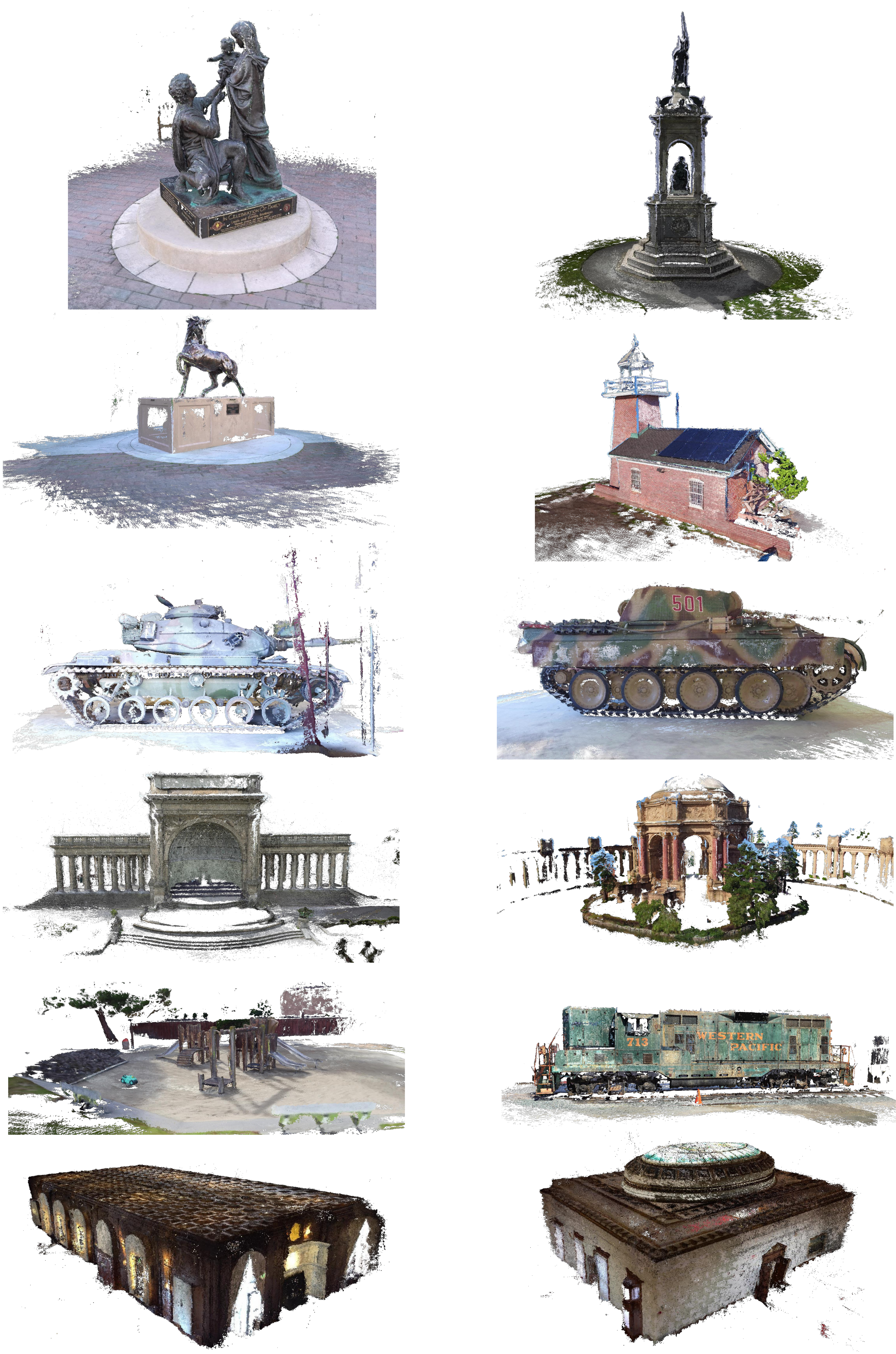}
\caption{Point cloud results on Tanks and Temples benchmark~\cite{knapitsch2017tanks}.}
\label{fig:tnt-pcd}
\vspace{-8pt}
\end{figure}

We visualize more point cloud results of KD-MVS (applied with CasMVSNet~\cite{gu2020cascade}) on DTU evaluation set~\cite{aanaes2016large}, Tanks and Temples benchmark~\cite{knapitsch2017tanks} respectively in \cref{fig:dtu-pcd} and \cref{fig:tnt-pcd}.

\section{Failure Cases}
As discussed in Sec. 5.3 of the paper, KD-MVS may face challenges when training student models under the following situations:
\vspace{-5mm}
\subsubsection{(a) When KD-MVS is applied on a relatively small-scale dataset.} We attempted to generate the pseudo probability distribution on the intermediate set of the Tanks and Temples dataset~\cite{knapitsch2017tanks} ($\sim 2K$ samples) for training student models. However, when trained on the dataset alone, the performance of the student model degrades significantly compared to the model trained 
on BlendedMVS dataset~\cite{yao2020blendedmvs} ($\sim 17K$ samples) alone. The potential reason is that small-scale datasets cannot provide sufficient data diversity for knowledge distillation, so the student model is not able to learn robust feature representations and performs unsatisfactorily.

\vspace{-5mm}
\subsubsection{(b) When the thresholds in cross-view check are inappropriate.} The performance of the student model relies on the quality of the generated pseudo labels, which is greatly affected by the thresholds set in cross-view check. \cref{tab:thershold} shows when these thresholds are inappropriate, the performance of the student model will degrade.

\end{document}